# SALIENCY DETECTION BY FORWARD AND BACKWARD CUES IN DEEP-CNN

*Nevrez İmamoğlu[1], Chi Zhang[2], Wataru Shimoda[1], Yuming Fang[2], Boxin Shi[1]*

[1]National Institute of Advanced Industrial Science and Technology, Tokyo, Japan
[2]School of Information Technology, Jiangxi University of Finance and Economics, Nanchang, China

## ABSTRACT

As prior knowledge of objects or object features helps us make relations for similar objects on attentional tasks, pre-trained deep convolutional neural networks (CNNs) can be used to detect salient objects on images regardless of the object class is in the network knowledge or not. In this paper, we propose a top-down saliency model using CNN, a weakly supervised CNN model trained for 1000 object labelling task from RGB images. The model detects attentive regions based on their objectness scores predicted by selected features from CNNs. To estimate the salient objects effectively, we combine both forward and backward features, while demonstrating that partially-guided backpropagation will provide sufficient information for selecting the features from forward run of CNN model. Finally, these top-down cues are enhanced with a state-of-the-art bottom-up model as complementing the overall saliency. As the proposed model is an effective integration of forward and backward cues through objectness without any supervision or regression to ground truth data, it gives promising results compared to state-of-the-art models in two different datasets.

## 1. INTRODUCTION

Visual attention mechanism helps us to investigate the visual scene by leading our attention to the significant regions [1-3]. This mechanism in the Human Visual System (HVS) inspires researchers to develop unsupervised or supervised algorithms to compute saliency maps for various computer vision applications [3-6].

With the recent trends using Deep Learning models such as Convolutional Neural Networks (CNNs) [7], it is now possible to recognize many objects [8] or to segment these objects [9] in complex scenes with high accuracy on color images. As demonstrated in recent works of CNN based top-down saliency models, the feature representations of the scene by learned filters in these complex networks are reliable cues for computational selective attention process. For example, CNN features can be used to make models to regress these features for inferring salient regions [10-12]. So, in addition to the CNN model learning, a secondary supervision on CNN features to the pixel level ground truth saliency maps can be done.

On the other hand, some studies [13, 14] have demonstrated that weakly supervised CNN models for object classification tasks can also be used to compute attentive regions for salient object detection. These models [13-14] suggest that for each class predictions on a CNN model, which is trained for recognizing 1000 objects on the scene, back-propagation (BP) [13] or guided-back-propagation (GBP) [14] can help to obtain objectness based distinct-class saliency maps as top-down saliency approaches.

In GBP [14], negative values are set to zero in all Rectified Linear Unit (ReLU) layers during backward process that enable improved class dependent salient regions. Compared to [10-12], objectness based methods using backpropagation do not require an end-to-end learning (e.g. regression of features to ground truth saliency maps). In both BP and GBP based saliency detection models, they calculate saliency maps from objectness for each class/object detected on the scene separately by assigning initial gradient as 1 for the relevant object and 0 for the others.

Fully supervised models using ground truth saliency maps such as [11] and [12] have been yielding good results on salient object detection. However, it is costly to prepare a training data with large number of ground truth data with pixel-level salient objects or regions labeled as binary images relevant to input images.

In addition, regarding the weakly supervised approaches in [13] and [14], these approaches give distinct class saliency maps for each object defined in the output of CNN classification task. So, they [13-14] are also costly for general salient object detection task. And, activations of CNN features, which may carry salient cues, are omitted in these works with backward process.

In this study, we build a top-down saliency detection model using the VGG-16 model by Simonyan *et. al.* [15, 13] trained for ImageNet dataset for 1000 objects. To extract salient features dependent on the objectness scores, we combine forward and backward salient cues, while demonstrating that partially guided backpropagation will provide sufficient information for selecting the features from forward run of CNN model. Therefore, salient object detection can be achieved with high accuracy whether the objects in the

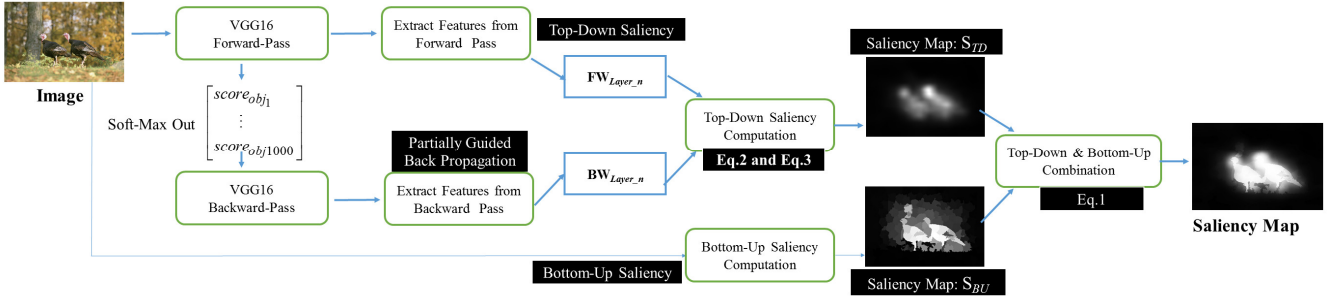
Figure 1 Flow of the proposed saliency computation model

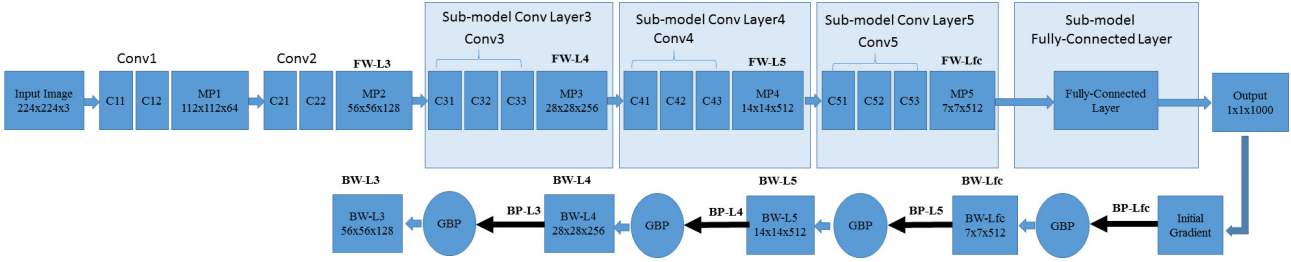
Figure 2 VGG16 CNN model with forward and backward processes

images are part of the object classes used for the training of CNN or not. We assume that object similarities enable the proposed model to find out salient regions even though object scores are low for the given test image. While the proposed model integrates forward and backward cues through objectness without any supervision or regression to ground truth data, it gives promising results compared to state-of-the-art models in two different datasets.

## 2. PROPOSED SALIENCY DETECTION MODEL

Our saliency detection model is inspired by weakly supervised models, which are taking advantage of back-propagation as in [13, 14]. The contributions of the proposed method with the main differences compared to GBP based model in [14] can be given as follows: 1) During back propagation, in addition to using 0 or 1 for initial gradients at the output level, we use the actual score values (object class predictions) obtained from the soft-max output of CNN to define attention values of regions based on their objectness scores. 2) By using score values for all objects at once, we do not separate BP for each class distinctly, and thus, the saliency map could be computed at one process. 3) Rather than utilizing fully guided back-propagation, we used partially guided back-propagation between separated sub-modules with each max-pooling layer for simplicity and efficiency. 4) We use salient cues both from forward and backward computations of the CNN model for selecting features for objectness based selective attention process for the given input images. 5) We modulate the top-down salient cues with bottom-up salient cues from a state-of-the-art model for enhancing the overall saliency.

In the proposed saliency model (Fig.1), we use VGG16 CNN [15, 13] trained to classify and label 1000 objects on images (using ImageNet dataset [16]). From VGG16, forward and backward cues are extracted from various layers, and combined linearly. Then, as a final step, a bottom-up saliency model modulates top-down cues to enhance the overall saliency map for the given input image. The sub-models used for obtaining salient cues from forward and backward process are three convolution layers (see Fig.2): Conv3, Conv4, Conv5, and one fully connected layer (fc). And, each of the Convolution sub-models includes three layers of convolution followed by three ReLU operations, and a final max-pooling for down-sampling as we think these blocks as a sequence of operation within a same context scale. Saliency map computation can be formulated as in Eq.1-3.

$$\mathbf{S} = \mathbf{S}_{TD} \times e^{\mathbf{S}_{BU}} \quad (1)$$

$$\mathbf{S}_{TD} = \sum_{n=\{3,4,5,fc\}} \left( W_n \times \mathbf{M} \times \mathbf{S}_{Layer\_n} \right) \quad (2)$$

$$\mathbf{S}_{Layer\_n} = \sum_{axis=features} \mathbf{FW}_{Layer\_n} \times \mathbf{BW}_{Layer\_n} \quad (3)$$

In Eq.1, $\mathbf{S}$ denotes the saliency map calculated for the given input image; $\mathbf{S}_{BU}$ is the bottom-up saliency map based on [17], and $\mathbf{S}_{TD}$ is the proposed top-down saliency integrating forward and backward cues. Exponential function is selected in Eq.1 because we want to keep $\mathbf{S}_{TU}$ unchanged if $\mathbf{S}_{BU}$ value is zero at a pixel, and we want only to amplify $\mathbf{S}_{TU}$ without reducing it. In Eq.2, $n$ is the sub-models {conv3, conv4, conv5, fc}, $\mathbf{S}_{Layer\_n}$ is the feature maps obtained from the sub-models in VGG16; $W_n$ represents the weights corresponding to each feature map; center-bias weighting $\mathbf{M} = 1 - $ *distance to center*, and scaled to {0.25 - 1} with size of 224x224. Because CNN models perform poorly around the borders (e.g. corners) of the input data. In Eq.2, the weighting values of $W_n$ are set to {1,5,10,1}. The weighting values are selected empirically after observing the saliency cues obtained from Eq.2. Eq.3 denotes the integration of forward ($\mathbf{FW}_{Layer\_n}$) and backward ($\mathbf{BW}_{Layer\_n}$) saliency cues from sub-model layers.

From each sub-model to get backward salient cues, we use back-propagation [13, 14] on output (class scores as the initial gradient) with respect to input features of corresponding layer. Distinct class saliency models as in [13] and [14] assign all initial gradients as zeros except the specific class of interest assigned as one during the back-propagation process. On the other hand, we use all the real class scores as initial gradient, so that the objectness level of each class could be expressed in the salient cues. In addition, this modification enables us to obtain salient cues for all possible objects at one time backward pass rather than repeating this operation for each object separately.

Back propagation is utilized for the backward process. However, we apply partially guided back-propagation by removing negative values to calculate $BW_{Layer\_n}$ object dependent backward salient cues. While calculating $BW_{Layer\_n-1}$, we only apply guiding process partially while passing the derivative results of sub-model $BW_{Layer\_n}$ to sub-model $BW_{Layer\_n-1}$. However, in original fully guided back-propagation (GBP), negative values set to zero at every ReLU layers. This partially guided back-propagation provides better results than using only back-propagation without guiding process. And, it is more efficient with saliency results like applying fully guided-back propagation. Guided back propagation yields sparse salient regions without keeping the texture connected on the object even though it has reliable accuracy of salient region detection. On the other hand, partial guiding process just between sub-models results in intact textures that improves the overall salient object detection task.

It should be noted that all the matrix multiplications in Eq.1 to Eq.3 are element-wise operations. Since the input data accepted by the VGG16 model is 224x224x3 (rows, columns, and color channels), we initially resize row and column of the feature maps, $S_{Layer\_n}$, to 224x224. Then, we perform Gaussian blurring on each $S_{Layer\_n}$ before combining them as in Eq.1. The kernel size of the smoothing filter is selected as 11x11 with $\sigma_X$ and $\sigma_Y$ defined as zero. Then, as in Eq.3, by using element wise multiplication, we select the salient features for each sub-model by the integration of forward and backward features over the feature channels. And, as stated before, all the feature maps from these sub-models are combined linearly (Eq.2). It is well accepted that there are two mechanisms in visual attention; bottom-up and top-down processes [2, 3]. Therefore, as a remaining element of the proposed model, a bottom-up model, $S_{BU}$, is used to modulate salient cues (Eq.1) obtained by the described top-down objectness based approach. Here, the work in [17] is used as bottom-up saliency model, which is designed based on graph-based manifold ranking (MR). MR [17] is a recent work and can obtain promising results in saliency prediction. We will demonstrate that combining these two different salient cues as in Eq.1 improves the overall performance of saliency detection. In the final step, we apply normalization operation to calculate the final saliency map as in Eq.4.

$$\text{Saliency} = \frac{1}{1+e^{-\eta \times (S-\text{mean}(S))}} \qquad (4)$$

## 3. EXPERIMENTAL RESULTS

In this section, we conduct the comparison experiments to demonstrate the performance of the proposed method by using Area Under Curve (AUC) metric obtained by the Receiver Operating Characteristic curve [18, 19]. Two publicly available datasets are selected for evaluation due their relative complexity compared to commonly used datasets. All images from these two datasets are also combined in the experiments, which result in a collection of 2447 images with different complexity. These datasets are: i) **ECCV2014-RGBD** [20]: this dataset includes 1000 images, and many of these images have single objects but they have more complex scenes and low color contrast in other images. It is collected in the study [20] and used to test RGB-D saliency. ii) **HKU-IS** [12]: this dataset has 1447 test images among overall 4447 images with various complexity in the scene such as multiple-objects, lower-contrast, ant etc. We use only test data of this dataset for our evaluation.

The following RGB based bottom-up and top-down saliency detection models are selected for the comparison: **GS** [21], **MR** [17], **SF** [22], **PCA** [23], **Y2D** [19], **RBD** [24], and **MDF** [12]. We also test the top-down saliency obtained by using back-propagation process on CNN model which can provide the object class scores with respect to input data (images). So, this back-propagation based saliency (**BPS**) is a variation of object class saliency proposed by Simonyan *et al.* [13, 15]. However, rather than calculating saliency for each possible object separately, as in our model, object class score vector is used as initial gradient for the process. Also, for the **BPS** [13, 15], results are smoothed with Gaussian filter to enhance the result during comparison. Moreover, we also apply fully guided back propagation (**FG**) in [14] to our model (Fig.1) to compare the performance with the proposed partially guided back-propagation (**PG**) based saliency detection. And, we refer to the overall proposed saliency with bottom-up saliency enhancement as the proposed **PGM** model. Some saliency samples of the proposed model and selected-state-of-the-art models are given in Fig.3.

The AUC results from the compared models for these two datasets are given in Fig.4, respectively. Initially, the evaluation is done on ECCV2014-RGBD dataset [20]. The proposed PGM model performs best on this dataset with AUC 0.9515, and the second-best result is from MDF [12] with AUC 0.9328. While MR [17] and our PG without bottom-up enhancement can obtain similar performance, and MR [17] is the third best model with AUC 0.9283. However, as it can be seen, even though both our PG model and MR [17] perform relatively worse compared with MDF [12] model, our PGM, enhancing PG with MR [23], outperforms other models with significant performance improvement. These initial results demonstrate that top-down and bottom-up models can be complementary saliency cues to obtain improved results.

On the other hand, as given in Fig.4, for the HKU-IS dataset, MDF [12] has the best performance with AUC value 0.9755. The proposed PGM model performs the second best with AUC 0.9569. And, our PG and FG without bottom-up salient cue enhancement is the third and fourth best performing result with enhancement is the third and fourth best performing result with AUC 0.9426 and 0.9299, respectively. It is interesting that there is a drastic performance difference for MDF [12] in two different datasets (more than 0.04 as 0.9328 and 0.9755 respectively). Probably, the reason is that HKU-IS test and train images include similar properties for MDF [12] to perform better. And, the images in ECCV-2014 dataset might have different properties compared to HKU-IS images. However, the proposed PG and PGM methods have very stable performances, especially in case of PGM in both datasets with AUC values 0.9515 and 0.9569. So, the proposed model gets consistent results for various datasets with variable image properties. Performance ranking for the HKU-IS dataset is also similar for the average overall AUC values as given in Fig.4. In addition, computation time (CPU only) of these models averaged from randomly selected images is also included in Table 1 (Test PC spec: Intel i5 6500, 32 GB DDR3 RAM). The proposed top-down saliency model using partially guided back propagation with or without bottom-up saliency enhancement outperforms other models except MDF [12]. It is important to note that the proposed saliency with PG performs better than that of using BPS [13] and fully-guided back propagation (FG) used in [14]. MDF [12] performs the best for all 2447 images; however, it is the slowest model as average processing cost for an image is around five minutes (Table 1). However, despite being slower than some of the bottom-up approaches due to its deep computational structure, the proposed model requires approximately two seconds for CPU only computation with high accuracy. The results demonstrate that the proposed method proves its reliability by providing consistent saliency maps with high accuracy and relatively low computation time.

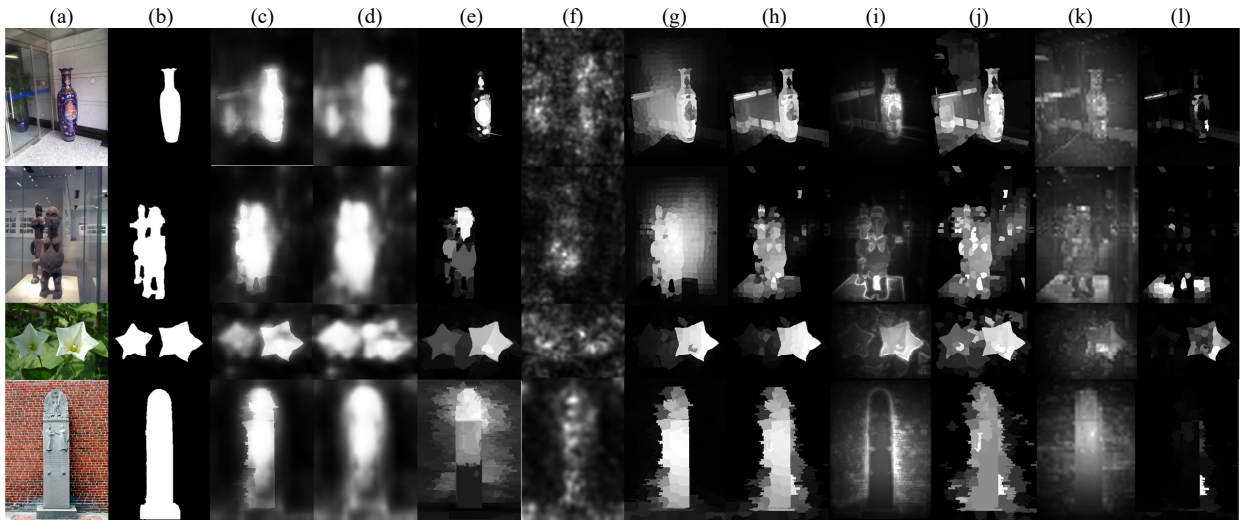

Figure 3 Sample color images (a) with their ground truth salient objects (b), and their corresponding saliency results for the proposed and state-of-the-art-models: c) Our PGM, d) Our PG, e) MDF [12], f) BPS [13], g) MR [17], h) RBD [24], i) PCA [23], j) GS [21], k) Y2D [19], l) SF [22]

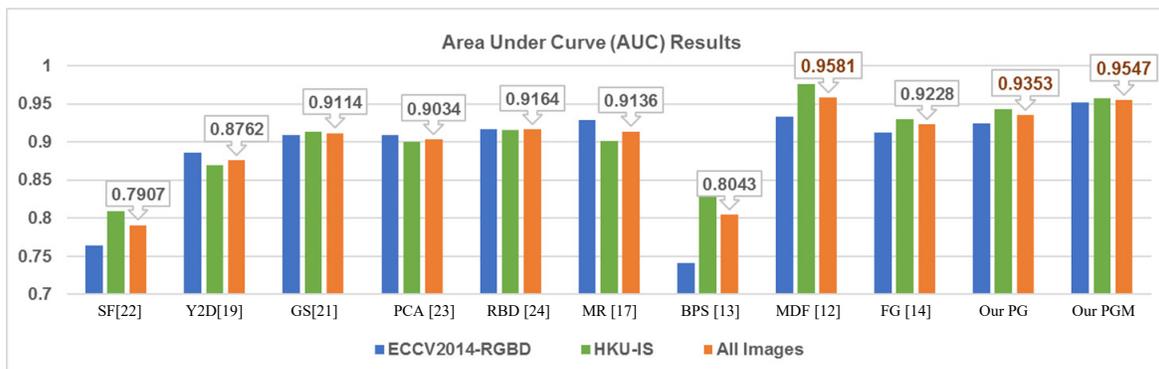

Figure 4 Evaluation Results of the State-of-the-Art Methods and Proposed Model Using Area Under Curve (AUC)

Table 1 Comparison of computation times (seconds) for selected models (CPU only performances)

| SF[22] | Y2D[19] | GS[21] | PCA [23] | RBD [24] | MR [17] | BPS [13] | MDF [12] | FG [14] | PG | PGM |
|---|---|---|---|---|---|---|---|---|---|---|
| 0.15 | 2.14 | 0.15 | 2.57 | 0.16 | 0.38 | 1.43 | 290.20 | 1.49 | 1.62 | 2.00 |

## 4. CONCLUSION

We introduced a saliency model that utilizes features from both forward and backward processes of CNN using the object scores. And, for backpropagation, we demonstrated that partial guiding (**PG**) yields better results by giving more uniform areas compared to fully-guiding (**FG**) used in [14]. Moreover, top-down saliency cues based on objectness scores can be enhanced by bottom-up salient features.

## 5. ACKNOWLEDGEMENTS